\documentclass[letterpaper, 10 pt, conference]{ieee_template/ieeeconf}  
\usepackage{amsmath}
\usepackage{amssymb}
\usepackage{booktabs}
\usepackage{caption}
\usepackage{epsfig}
\usepackage{graphicx}
\usepackage{multirow}
\usepackage{pifont}
\usepackage[table]{xcolor}
\usepackage{xspace}
\usepackage[percent]{overpic}
\usepackage{float}
\usepackage{xfrac}

\definecolor{darkblue}{RGB}{0, 0, 200}
\newcommand{\xmb}{{\color{darkblue} \ding{55}}}

\makeatletter
\let\NAT@parse\undefined
\DeclareRobustCommand\onedot{\futurelet\@let@token\@onedot}
\def\@onedot{\ifx\@let@token.\else.\null\fi\xspace}
 
\def\ie{\emph{i.e}\onedot} 
\def\etal{\emph{et al}\onedot}
\g@addto@macro\@maketitle{
\begin{figure}[H]
  \captionsetup{singlelinecheck=false, font=small, skip=4pt, belowskip=-6pt}
  \setlength{\hsize}{\textwidth}
   \vspace{-4mm}
  \centering
\begin{tabular}{ccc}
 	\begin{overpic}[width=0.325\textwidth, height=4cm,keepaspectratio]{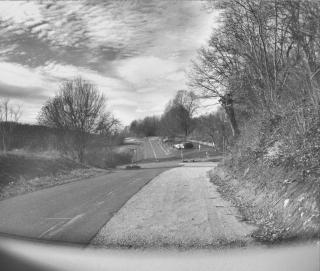}
    \put (0,3) {\colorbox{lightgray}{$\displaystyle\textcolor{green}{\text{(a)}}$}}
    \end{overpic}
    \begin{overpic}[width=0.325\textwidth, height=4cm,keepaspectratio]{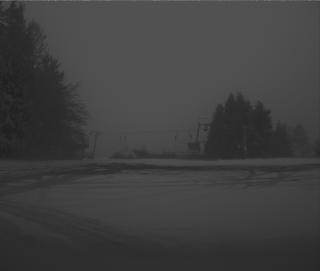}
    \put (0,3) {\colorbox{lightgray}{$\displaystyle\textcolor{green}{\text{(b)}}$}}
    \end{overpic}
    \begin{overpic}[width=0.325\textwidth, height=4cm,keepaspectratio]{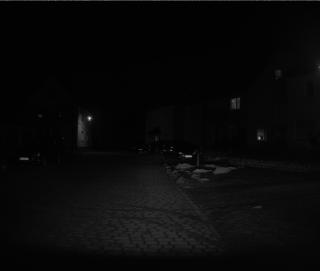}
    \put (0,3) {\colorbox{lightgray}{$\displaystyle\textcolor{green}{\text{(c)}}$}}
    \end{overpic} \\

 	\begin{overpic}[width=0.325\textwidth, height=4cm,keepaspectratio]{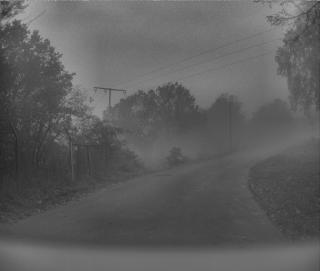}
    \put (0,3) {\colorbox{lightgray}{$\displaystyle\textcolor{green}{\text{(d)}}$}}
    \end{overpic} 
    \begin{overpic}[width=0.325\textwidth, height=4cm,keepaspectratio]{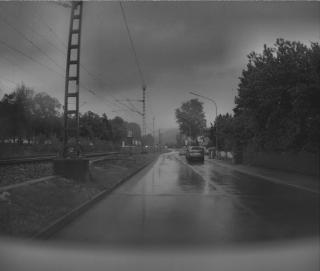}
    \put (0,3) {\colorbox{lightgray}{$\displaystyle\textcolor{green}{\text{(e)}}$}}
    \end{overpic} 
    \begin{overpic}[width=0.325\textwidth, height=4cm,keepaspectratio]{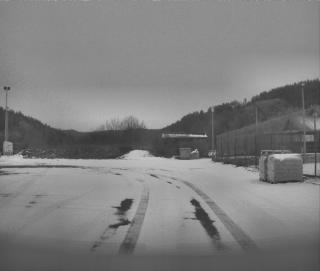}
    \put (0,3) {\colorbox{lightgray}{$\displaystyle\textcolor{green}{\text{(f)}}$}}
    \end{overpic}
\end{tabular}
\caption{\textbf{Sample images from our dataset.} (a) Bright, (b) Moderate Light, (c) Low Light, (d) Fog, (e) Rain and (f) Snow.}
\label{sampleimg}
\end{figure}
\vspace{-1.25em}
}
\makeatother

\usepackage[colorlinks,pagebackref=false,citecolor=blue,bookmarks=false,hypertexnames=true]{hyperref}
\begin{document}

\title{Weather and Light Level Classification for Autonomous Driving:\\ Dataset, Baseline and Active Learning}

\author{
Mahesh M Dhananjaya$^{1}$,
Varun Ravi Kumar$^{1}$ and
Senthil Yogamani$^{2}$ \\
{\normalsize 
$^{1}$Valeo DAR Kronach, Germany \quad
$^{2}$Valeo Vision Systems, Ireland}
}

\maketitle
\begin{abstract}
Autonomous driving is rapidly advancing, and Level 2 functions are becoming a standard feature. One of the foremost outstanding hurdles is to obtain robust visual perception in harsh weather and low light conditions where accuracy degradation is severe. It is critical to have a weather classification model to decrease visual perception confidence during these scenarios. Thus, we have built a new dataset for weather (fog, rain, and snow) classification and light level (bright, moderate, and low) classification. Furthermore, we provide street type (asphalt, grass, and cobblestone) classification, leading to 9 labels. Each image has three labels corresponding to weather, light level, and street type. We recorded the data utilizing an industrial front camera of RCCC (red/clear) format with a resolution of $1024\times1084$. We collected 15k video sequences and sampled 60k images. We implement an active learning framework to reduce the dataset's redundancy and find the optimal set of frames for training a model. We distilled the 60k images further to 1.1k images, which will be shared publicly after privacy anonymization. There is no public dataset for weather and light level classification focused on autonomous driving to the best of our knowledge. The baseline ResNet18 network used for weather classification achieves state-of-the-art results in two non-automotive weather classification public datasets but significantly lower accuracy on our proposed dataset, demonstrating it is not saturated and needs further research.
\end{abstract}
\section{Introduction}

Detecting weather conditions remains one of the most difficult problems in designing secure and effective advanced driver assistance systems (ADAS). 
Recognition of weather and visual conditions plays a significant part in different sensors' ability and poses many risks in developing an autonomous car with changing weather and visual conditions. Rain can degrade the performance of commonly used automotive sensors like LiDAR  \cite{goodin2019predicting}. It is critical to detect rain and snow when LiDAR performs poorly. 
Compared to driving in clear weather, driving in the rain/snow/low-light poses a higher risk \cite{rama2001effects}. Road safety analysis should include the system behavior in these challenging conditions. For example, object detection algorithms must perform well in low light and snowy weather, as well as clear conditions. In practice, the performance degrades drastically in these conditions. Alternatively, a specialized model for these conditions can be designed.


In this article, visual conditions are defined as a noticeable variation in brightness as bright, moderate, or low light across dawn/dusk, day and night. In contrast, weather conditions are precipitation changes in the environment, including clear weather, rainy weather, and snowy weather. Instead of letting a network predict one class, it is also possible to train a network to predict several classes at once, \ie, visual conditions and weather conditions. Such a multi-task \cite{roh2019survey} approach takes advantage of the effective use of parameters in a deep model. It facilitates feature sharing between classes that can impact the overall performance. In an autonomous driving system, weather classification can aid the removal of lens soiling (de-soiling) \cite{uvrivcavr2019soilingnet, uricar2019desoiling}, and light level classification can aid ISP tuning \cite{yahiaoui2019overview}.\par

Active learning, a form of semi-supervised learning in which the algorithm selects the data it needs to learn, is used along with transfer learning, which helps in weights initialization and feature transfer from one type of dataset to another. With this approach, the model will actively question an authority source, either the software engineer or a labeled dataset, to be told the correct prediction for a given problem. To realize this scenario, a small trainable module, named "\textit{loss prediction module}" inspired by Donggeun Yoo \etal \cite{yoo2019learning} is attached to the deep neural network model. 
This loss prediction module is trained by reducing one common loss, irrespective of the tasks and architecture complexity. If the loss of a data point can be estimated, significant losses are likely to occur in data points for wrong predictions. These data points selected would provide the current model with additional information. The model can be used to label the rest of unlabeled and new data with suitable thresholding in the prediction's confidence level  \cite{roh2019survey}. It leads to a high precision model; most data can be auto-labeled, and the rest few can be unlabelled or sent to human labeling. The advantage of this is that it reduces labeling cost, and simultaneously a better model is evolved. It helps to keep the consistency of labeling and visualize the features being considered since uncertainty in corner cases are the cause for errors even in determining ground-truth labels from human beings' perspective as the parameter used may be too rigid to measure \cite{mccartney1976optics}.\par

Implementing such an architecture, we can summarize that a diverse and unconstrained dataset perceived by autonomous driving cameras with weather and visual conditions annotations can be aggregated. Using this dataset and the loss prediction module that selects data, we introduce a quick and effective active learning process. Thus, we provide a framework to auto-label new data with minimum manual labelling.\par
\section{Related Work}

\subsection{Weather Classification}
Scene awareness regarding weather and visual conditions must be considered a multi-label classification task since they may appear simultaneously, and feature sharing between domains can impact the overall performance. Previously the two domains were studied individually or as a binary classification task, \ie, belongs to a specific label or not. Deep neural network models outperform all of the previous nominal methods related to Mathematical models, filter-based models, and machine learning models using shallow networks, as discussed by Ibrahim \etal \cite{ibrahim2019weathernet}.\par 

Guerra \etal \cite{guerra2018weather} developed a framework to detect rain, fog, and snow using super-pixel masks in Convolutional Neural Network and Support Vector Machine classifiers. It only considers that the above conditions are exclusive to each other, ignoring multiple classes' existence on a single image. Zhao \etal \cite{zhao2018cnn} addressed that multiple labels are in a single image and uses CNN to extract the preliminary features and convolutional LSTM to predict weather labels. While this model demonstrates improvement in the examination of various weather conditions (cloudy, snowy, rainy, foggy, and sunny), but overlooked the influence of light conditions.\par

The WeatherNet \cite{ibrahim2019weathernet}  consists of four deep CNN models to detect (dawn/dusk, day, night-time), (glare), (rain, snow), and (fog) respectively. This approach addressed the exclusivity and co-existence of classification labels. However, the use of the CNN model for each class made the framework computationally heavy and neglected feature sharing. Dawei Xuan \etal \cite{xia2020resnet15} ResNet15 based model for weather recognition on traffic road simplified and improved residual network structure from ResNet50. It demonstrated that the usage of smaller networks is suitable to extract weather characteristics from Traffic Road images. However, this is classified as sunny, rainy, snowy, and foggy, limiting only to severe weather conditions and ignoring other visual conditions' influence. A change in architecture meant that it had to be trained from random initialization.\par
\setcounter{figure}{1}
\begin{figure*}[!t]
\captionsetup{singlelinecheck=false, font=small, skip=2pt, belowskip=-8pt}
\centering
\includegraphics[width=\textwidth]{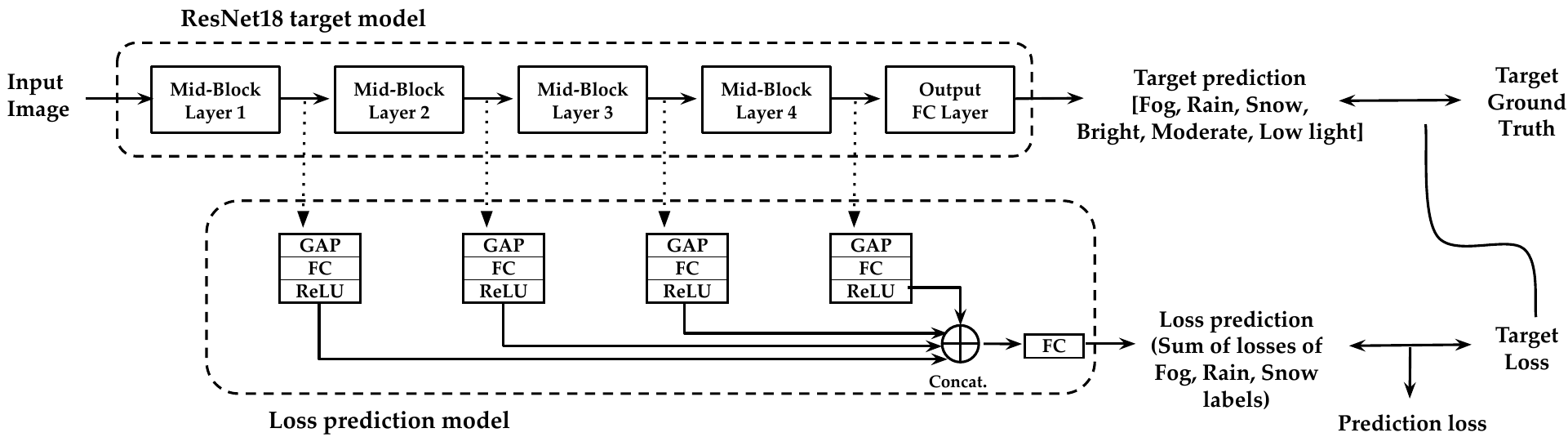}
\caption{\textbf{Illustration of the loss prediction model for weather and light level classification using the framework proposed in \cite{yoo2019learning}.} It is attached to a number of intermediate layers of the target model, outputting an estimate of loss for any input image.}
\label{losspred2}
\end{figure*}
\subsection{Active Learning}
The main principle behind active learning is allowing a machine learning algorithm to prefer the data from which it learns to allow greater precision with fewer training labels. Active learning is inspired by various modern machine problems where unlabelled data can be readily available. Labels, however, difficult to procure, time-consuming, and costly \cite{settles2009active}. Active learning reduces the time and cost required to label by a human annotator.\par

The selection of data to train from a unlabeled database is possible through three primary approaches: an uncertainty-based approach, a diversity-based approach, and expected model change. The uncertainty method  \cite{wang2016cost} describes and calculates the amount of uncertainty before selecting data with a high degree of uncertainty. The diversity strategy  \cite{sener2018active}, on the other hand, chooses random data points from the unlabeled database. Expected model change \cite{freytag2014selecting} chooses data that, if the prediction was known, would cause the most change to the existing model parameters and the overall performance.\par

To describe uncertainty, the most basic form of the uncertainty strategy is to use class posterior probabilities. An entropy of class posterior probabilities \cite{joshi2009multi} or the probability of a predicted class \cite{wang2016cost} defines the uncertainty of a data point. Despite its simplicity, the uncertainty method \cite{lewis1994sequential} for classification tasks always performs well. However, since network outputs are used, a task-specific architecture is needed. Donggeun \etal \cite{yoo2019learning} designed a loss prediction model that comes under the uncertainty method but differs in that it forecasts "loss" based on the backbone model's intermediate features than statistically predicting uncertainty from predicted outputs alone. A loss prediction model aims to find a hard example that represents high-loss training data \cite{shrivastava2016training} by forecasting these data using only intermediate features, without the need for annotations of data.\par
\section{Dataset and Baseline}

\subsection{Proposed Dataset}

We use an industrial front camera mounted on the top of the windshield of a car. The camera outputs a resolution of $1024\times1084$ video stream in RCCC format. R refers to a red pixel and C refers to a clear pixel capturing intensity. 
RCCC format provides a higher resolution intensity than a Bayer pattern RGB image at the cost of reduced color information. 
The data was recorded over two years in Germany to capture different weather scenarios. It was also captured over the day, twilight, and night times to get various light levels. A total of 15k video clips were collected. From these 15k videos, 60k images were picked using stratified sampling techniques discussed in \cite{uricar2019challenges}. A subset of these samples was manually labeled with three categories, namely weather (clear, rain, and snow), light level (bright, moderate, and low), and street type (asphalt, grass, and cobblestone). The active learning framework bootstraps and auto-labels the remaining frames. Each image has three labels corresponding to a category, and overall there are nine possible labels. The dataset can also be effectively used to generate more samples using GAN techniques \cite{uricar2021let}.\par
\subsection{Baseline Classification Network}

The images are down-scaled for training from the captured resolution of $1024\times1084$ to $224\times224$ to match the ResNet18 ImageNet architecture. Usage of a standard backbone also enables easy integration with multi-task shared encoder models where this task can be added to existing perception model with incremental increase in complexity \cite{chennupati2019multinet++, kumar2021omnidet}.
This down-scaling had a minimal impact on the performance of the model. We only used grayscale values and ignored the red color information as our initial experiments found it to have a minor effect. As our work focuses on weather and light level classification, we have not used the street type classification labels. We use the ResNet18 \cite{he2016deep} classification model with pre-trained weights from ImageNet dataset \cite{deng2009imagenet} and modify it to classify light level and weather categories outputting 6 labels. Usage of a standard encoder also enables a straightforward integration into a multi-task perception network. In fine-tuning, the learning rate is kept low, and a custom rate scheduler was designed.\par
\subsection{Active Learning Training Framework} 
\label{activelearningsec}

\begin{figure*}[t]
\captionsetup{singlelinecheck=false, font=small, skip=2pt, belowskip=-10pt}
\centering
\includegraphics[width=1.6\columnwidth]{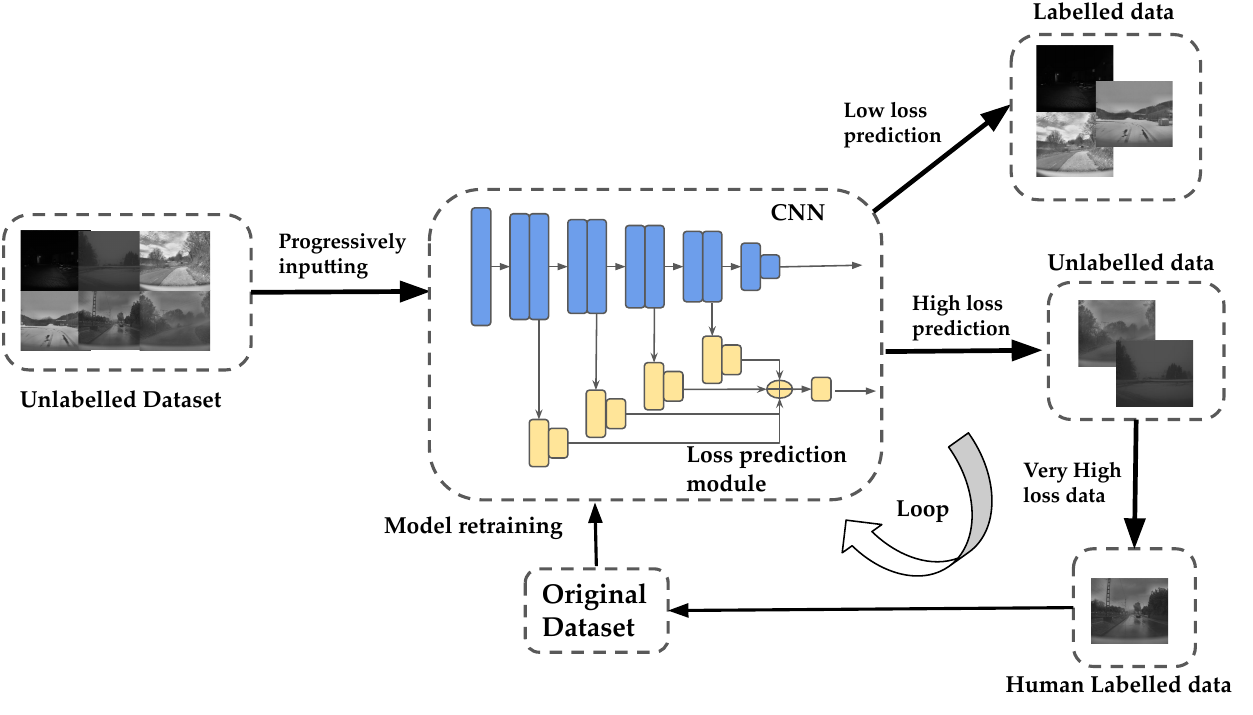}
\caption{\textbf{Overview of our active learning framework for weather and light level classification.}}
\label{overview}
\end{figure*}
\textbf{\textit{Motivation}}: In the supervised learning domain, we can improve deep neural networks' performance with additional data. The difficulty is that the cost for labeling is limited and often proven to be very expensive. We can not rely on having a supervised label for every possible image we encounter in the real world. One of the most prominent challenges for a real-world computer vision system is learning a representation from limited data that generalizes to novel situations. It is especially true for autonomous driving - the variety of road scenes that exist is substantial. There are many promising methods to learning robust representations. Active learning is a better option practically as obtaining accurate weather classification labels for every sample in a dataset is a labor-intensive process that costs significant amounts of money and time. In this work, we aim to use active learning for the following reasons:
\begin{itemize}
\item To make use of less amount of labeled data to achieve an equivalent accuracy using a random sampling of data. 
\item To detect images for which the model is uncertain or not confident in its classification \cite{yoo2019learning}. In the case of an auto-labeler, the data can be queried by the user. The prediction system can be employed to detect classification uncertainty and make use of other redundant systems.
\item To set up the model to deal with new incoming data and further improve its performance.
\end{itemize}

As mentioned earlier, selection of data such that they have maximum impact on the network is necessary. Donggeun \etal \cite{yoo2019learning} proposed an active learning framework wherein a helper module actively selects data points based on the predicted loss. To realize this, a \textit{"loss prediction module"} is coupled with the target model, and the model learns to estimate the loss simultaneously with the target model prediction, as seen in Figure \ref{losspred2}. The active learning module considers intermediate features from the target module, then the loss is calculated by concatenating these features and mapping them into a single value.\par
For a given input image, the target model predicts a weather (clear / rain/snow) and light level (bright/moderate/low) condition, and the loss prediction module estimates the loss for the above prediction (see Figure~\ref{losspred2}). In the learning phase, the ground truth and the prediction are compared to produce a target loss. This target loss is used as a ground truth loss for the loss module, computing the loss-prediction loss. The loss-prediction loss is further used to train the loss prediction module to predict the losses with the annotation of the data images. Finally, the loss-prediction module is then employed for active learning in selecting data from an unlabeled data pool that needs to be labeled for retraining.\par
\textbf{\textit{Active learning process:}} Initially, we need to label a small sub-set of the dataset manually. It provides the essential difference between the labels, and later the model is trained on them. There exist significant uncertainty and lower scores from the predictions. However, this run will serve the model to get an insight into the parameter space areas that demand to be labeled further. During the early stages of the training, we observed that the number of layers to be trained should be minimal and freeze the rest of the layers with pre-trained weights due to significantly fewer data. With the increase in the trainable data subsequently, the number of layers required to train can also be increased. Finally, when the training is complete, the network can predict the class on the remaining unlabelled data.\par

In our case, the Loss from the Loss prediction module is used to choose data from the unlabelled database. The data with high predicted losses are manually refined and used for retraining. This process can be iterated as shown in Figure~\ref{overview}, to improvise the model's accuracy. As the models improve over time, one can continue to optimize the labeling technique. After each retraining cycle, a large amount of unlabelled data are annotated with high confidence by only training on a subset of high loss labeled data. We observed that, at each cycle of training on new labeled data, the parameter initialization is preferred to be the pre-trained weights from the ImageNet dataset and not from the previous cycle of training since, in a few cycles, it would cause over-fitting of data without improving the performance of the model.\par
\textbf{\textit{Auto labeling:}} We expect high precision from the network employed to annotate the unlabelled data, \ie, a high proportion of the identifications are correct. The predictions are divided into labeled, unlabelled, and data required to be labeled by humans as shown in Figure~\ref{overview}. We can achieve this by suitable thresholding the output of the loss prediction network \cite{yoo2019learning}. A small part of a very high loss can be queried to the user for manual labeling. A segment of the data with low loss is labeled, and the rest of the data can be classed as unlabeled data and relabelled with higher confidence in further learning process cycles.\par
\section{Results}

\begin{figure}[t]
\captionsetup{singlelinecheck=false, font=small, skip=2pt, belowskip=-10pt}
\centering
\includegraphics[trim=0 0 0 0,clip,width=\columnwidth]{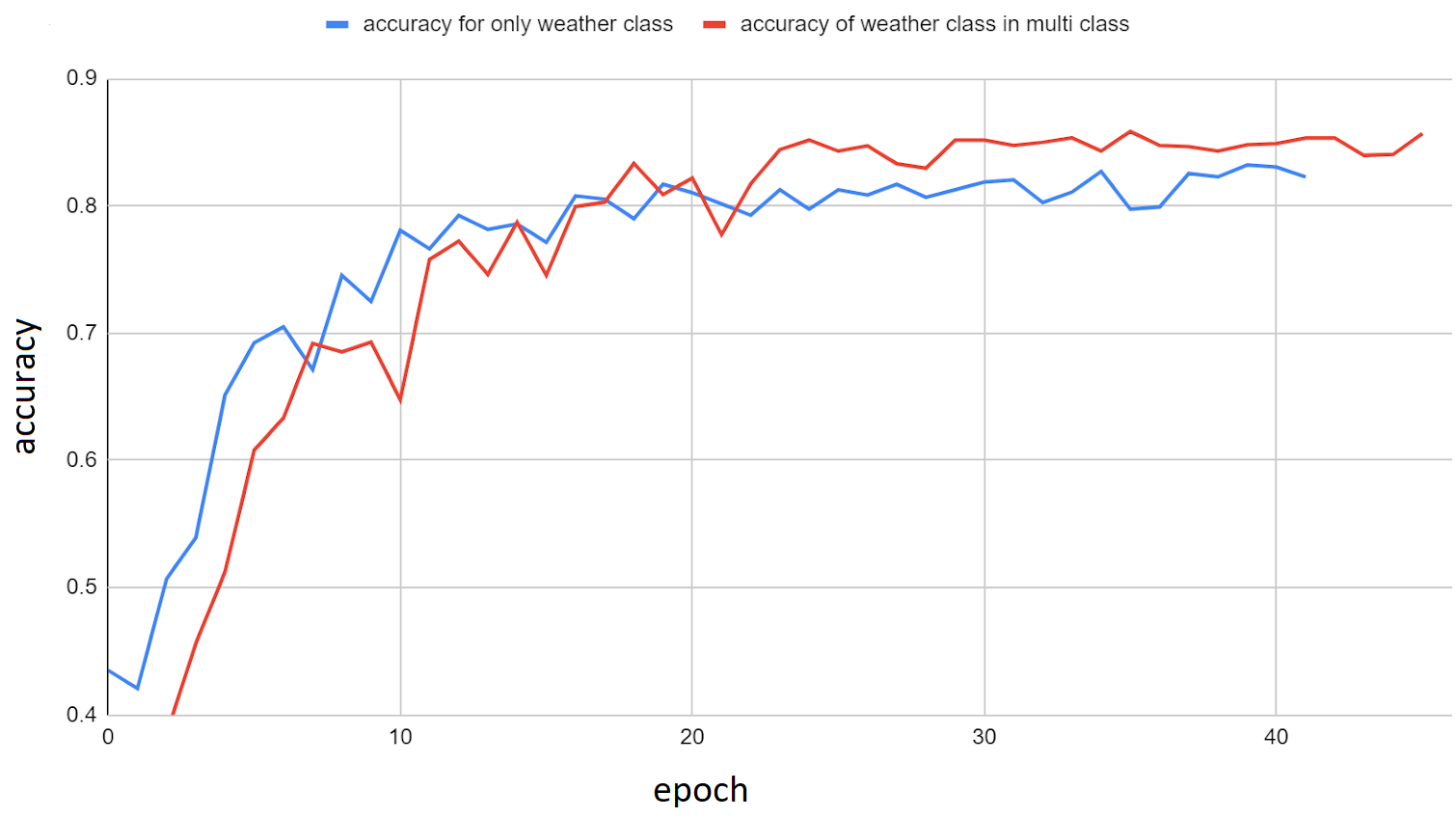}
\caption{\textbf{Comparison of a single class and multi-class training.} Multi-class training achieves better performance.}
\label{fig:multi-class}
\end{figure}

\begin{table*}[t]
\centering
\begin{tabular}{lcccccccccc}
\toprule
\textbf{Approach} &
  \multicolumn{1}{c}{\textit{\textbf{Dataset}}} &  \multicolumn{1}{c}{\textit{\textbf{\#Classes}}} & \multicolumn{1}{c}{\textit{\textbf{\#Images}}}  &
  \begin{tabular}[c]{@{}c@{}}\textit{\textbf{Overall}} \end{tabular} &
  \textit{\textbf{Sunny}} &
  \textit{\textbf{Cloudy}} &
  \textit{\textbf{Foggy}} &
  \textit{\textbf{Rainy}} &
  \textit{\textbf{Snowy}}  \\ 
  \midrule
Gbeminiyi \etal \cite{oluwafemi2019multi}  & \multirow{3}{*}{Multi-class \cite{gbeminiyi2018multi}} & \multirow{3}{*}{4} & \multirow{3}{*}{1125}  & \xmb & \xmb & \xmb & \xmb & 95.20$^*$ & \xmb  \\
WeatherNet~\cite{ibrahim2019weathernet}   & & &  & \xmb   & \xmb & \xmb & \xmb & 97.69$^*$ & \xmb \\
Ours & & &    & 98.80$^*$ & \xmb & \xmb & \xmb & \textbf{100}$^*$  & \xmb \\
\midrule
Zhao \etal\cite{zhao2018cnn}        & \multirow{3}{*}{Multi-Label \cite{zhao2018cnn}} & \multirow{3}{*}{5} & \multirow{3}{*}{10000} 
 & 0.870 & 0.8404 & \textbf{0.9346} & 0.8584 & 0.8040 & 0.9154 \\
WeatherNet~\cite{ibrahim2019weathernet} &  & &
& 0.853 &  \textbf{0.8728} & \xmb & \textbf{0.8832} & 0.7752 & 0.8820 \\
Ours & & & &  \textbf{0.872} & 0.8290 & 0.9185 & 0.8632 & \textbf{0.8147} & \textbf{0.9331} \\
\midrule
Ours & Proposed Dataset & 9 & 1165  & 0.772 & \xmb & \xmb & 0.7324 & 0.7510 & 0.7815 \\
\bottomrule
\end{tabular}
\caption{\textbf{Evaluation of our weather classification model on three datasets.} 
$F_1$ score is used as the metric except for the values marked with $^*$ which denotes accuracy. }
\label{tab:weather}
\end{table*}
To facilitate a comparative study, we apply our model to two open-sourced datasets used in previous studies by Zhao \etal\cite{zhao2018cnn}. Table~\ref{tab:weather} demonstrates the dataset's evaluation of our approach in comparison to the baseline  and WeatherNet approach \cite{ibrahim2019weathernet}. Our model was designed to have similar performance across all different labels and not prioritize a single weather condition. Thus the model could predict the less common weather conditions like rain and snow better than other models. Table \ref{tab:lightlevel} summarizes the baseline light level classification results. The model performs well in identifying bright and low light. As moderate light is semantically ambiguous, it has a significantly lesser $F_1$ score. Thus we demonstrate that the model identifies features across all labels through a better selection of data and model architecture. Figure \ref{fig:multi-class} shows that multi-class training achieves better performance than training the classes individually. This may be due to network learning better features for discriminating classes.\par
\begin{table}[t]
\centering
\begin{tabular}{lc}
\toprule
\textbf{Light Level} & \textbf{$F_1$ score} \\
\midrule
Low & 0.790 \\
Moderate & 0.711 \\
Bright & 0.862 \\
\bottomrule
\end{tabular}
\caption{Baseline results for light level classification.}
\label{tab:lightlevel}
\end{table}
\textbf{\textit{Transfer learning:}}
In machine learning, transfer learning is a fundamental approach for weights initialization to address the fundamental issue of inadequate training data. The weights are attempted to be transferred from the source domain to the target domain. In our work, we make use of network-based deep transfer learning \cite{tan2018survey}. It refers to reusing a pre-trained partial network (including network structure and parameter values) in the source domain and integrating it into a deep neural network in the target domain. The network's top layers are usually viewed as a function extractor, and the extracted features are modular. As an example, to start, the network is trained on a large-scale training dataset in the source domain (ImageNet). Secondly, this pre-trained partial network is migrated to a new network that has been configured for the target domain. Finally, we fine-tune our model by updating the network parameters. \cite{tan2018survey, roh2019survey}.

\begin{figure}[t]
\captionsetup{singlelinecheck=false, font=small, skip=2pt, belowskip=-10pt}
\centering
\includegraphics[trim=0 0 0 0,clip, width=0.8\columnwidth]{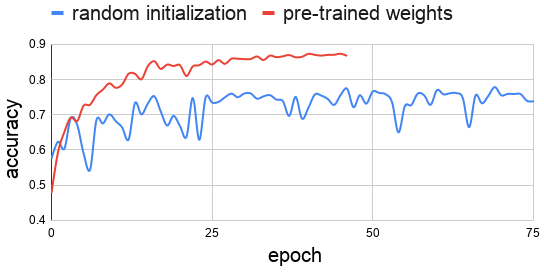}
\caption{\textbf{Comparison of random initialization and pre-trained weights.} Usage of pre-trained weights consistently performed better than random initialization.}
\label{fig:graph1}
\end{figure}

Figure \ref{fig:graph1} compares the accuracy between two models where the weights were randomly initialized (blue curve), and in the other, the pre-trained weights of Imagenet were used (red curve). It was noted that the learning process for the model with the use of pre-trained weights was faster and could attain considerable better accuracy over randomly initialized weights.\par
\begin{figure}[t]
\captionsetup{singlelinecheck=false, font=small, skip=2pt, belowskip=-10pt}
\centering
\includegraphics[width=0.8\columnwidth]{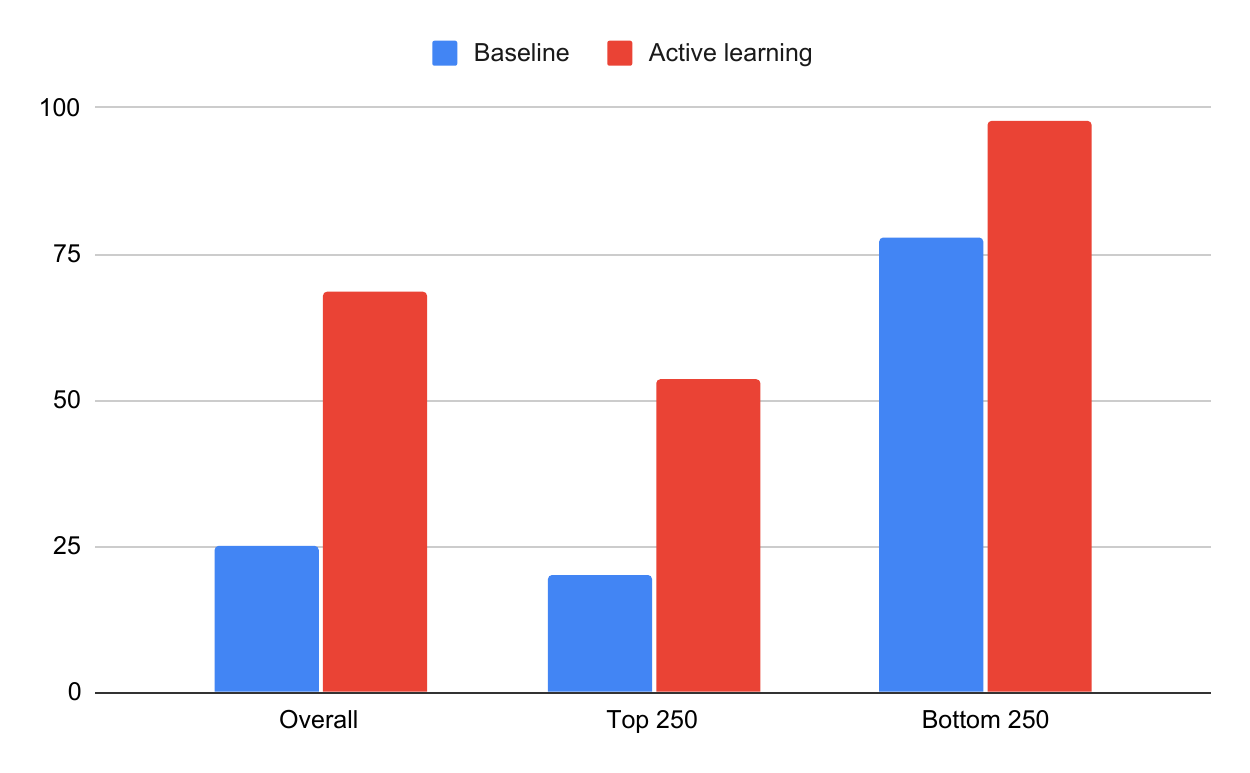}
\caption{\textbf{Loss prediction before and after active learning for snow labels.}}
\label{loss1}
\end{figure}

\textbf{\textit{Analysis of Loss prediction:}}
The loss value of an image indicates how well the model's prediction was on a single labeled sample. It can be used as a score to determine the images required to be trained. However, to calculate the loss, the image has to be labeled. In our case, there is a loss prediction module that predicts this loss using intermediate features. Higher the loss, higher the likelihood of the image having the wrong prediction, and lower the loss, higher the likelihood of the image having the correct prediction. Figure \ref{loss1} compares the accuracy for snow labels of top and bottom 250 images from a sorted list using loss prediction scores. Overall accuracy improved after one cycle of active learning, using the loss prediction module's images. In both cases, the top 250 image accuracy (relative higher loss) has a lower accuracy than the bottom 250 image accuracy (relative lower loss). The top k image accuracy provides the worst-case accuracy of the whole database, which helps determine the model's performance on an unlabelled data pool. In contrast, bottom k images are highly confident, optimistic predictions that can be auto-labeled.\par
\begin{figure}[t]
\captionsetup{singlelinecheck=false, font=small, skip=2pt, belowskip=-10pt}
\begin{center}
\includegraphics[trim=0 0 0 0cm,clip,width=\columnwidth]{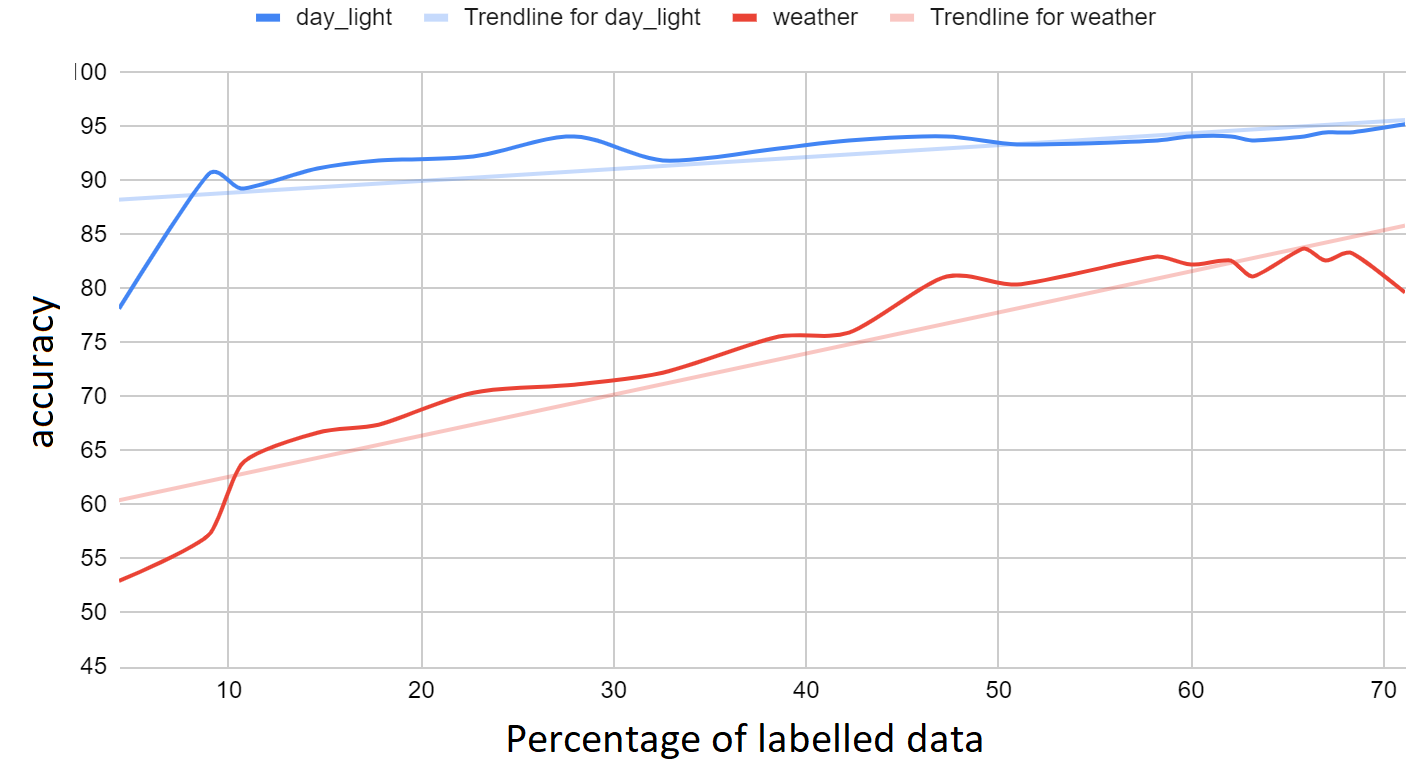}
\end{center}
\caption{\textbf{Data selection using active learning.}}
\label{fig:data_selection}
\end{figure}

\textbf{\textit{Data selection using active learning:}}
Initially, a random set of 500 images are selected from the 60k samples guided by stratified sampling \cite{uricar2019challenges}. These samples are manually labeled, and the baseline model is trained. The resulting model lacks high accuracy but sufficient to provide a primary classification. Later, we picked 200 images that have the highest predicted loss values among the remaining 60k samples. The labels of these new frames are manually refined, and the model is retrained. After a single cycle of active learning, there is considerable performance improvement, and further improvement is achieved by training multiple cycles. Figure \ref{fig:data_selection} illustrates that through initial cycles of this operation (up to 50\% of trainable data), it can be estimated that any new data obtained and selected through the loss prediction module will continually improve the accuracy of the model. At the highest accuracy, the model trained on fewer data selected through the loss prediction model can be approximately equivalent to training the model on all trainable data. It signifies that a subset of trainable data is required, with efficient data selection for an accurate model.\par
\section{Conclusion}
In this paper, we created a dataset for weather and light classification. It is crucial to identify these scenarios as visual perception degrades severely in low light and harsh weather conditions. We demonstrate our active learning implementation to distill a large dataset into 1.1k samples for a public release. The active learning framework can also automatically label large-scale object detection datasets to understand the impact of particular weather conditions and low light. We implemented a simple baseline model which achieves state-of-the-art results on two public non-automotive weather classification datasets. Relatively, accuracy on the proposed dataset is significantly lower, showing that the baseline does not saturate the dataset, and further design improvements are needed. In future work, we plan to explore the usage of weather classification to adapt perception models. \par
\bibliographystyle{ieee_template/IEEEtran}
\bibliography{bib/egbib}
\end{document}